\documentclass{article}
\input{taesup.sty}

\PassOptionsToPackage{numbers}{natbib}

\usepackage[final]{nips_2017}

\usepackage[utf8]{inputenc} 
\usepackage[T1]{fontenc}    
\usepackage{url}            
\usepackage{booktabs}       
\usepackage{amsfonts}       
\usepackage{nicefrac}       
\usepackage{microtype}      
\usepackage{multirow}
\usepackage{scalerel}
\usepackage[export]{adjustbox}
\usepackage{wrapfig}

\usepackage{natbib}

\usepackage{algorithm}
\usepackage{algorithmic}

\usepackage{dsfont}

\newcommand{\eg}{{\it e.g.}}
\newcommand{\ie}{{\it i.e.}}

\newcommand{\Lb}{\mathbf{\Lambda}}
\newcommand{\Ellb}{\mathbf{L}}
\newcommand{\E}{\mathbb{E}}

\title{Neural Affine Grayscale Image Denoising}

\author{
  Sungmin Cha, Taesup Moon
   \\
 College of Information and Communication Engineering\\ 
 Sungkyunkwan University, Suwon, Korea 16419 \\
  \texttt{tsmoon@skku.edu} \\
}

\begin{document}

\maketitle

\begin{abstract} 

We propose a new grayscale image denoiser, dubbed as Neural Affine Image Denoiser (Neural AIDE), which utilizes neural network in a novel way. Unlike other neural network based image denoising methods, which typically apply simple supervised learning to learn a mapping from a noisy patch to a clean patch, we formulate to train a neural network to learn an \emph{affine} mapping that gets applied to a noisy pixel, based on its context. Our formulation enables
both supervised training of the network from the labeled training dataset and adaptive fine-tuning of the network parameters using the given noisy image subject to denoising. The key tool for devising Neural AIDE is to devise an estimated loss function of the MSE of the affine mapping, solely based on the noisy data. 
As a result, our algorithm can outperform most of the recent state-of-the-art methods in the standard benchmark datasets. Moreover, our fine-tuning method can nicely overcome one of the drawbacks of the patch-level supervised learning methods in image denoising; namely, a supervised trained model with a mismatched noise variance can be mostly corrected as long as we have the matched noise variance during the fine-tuning step.




\end{abstract}


\section{Introduction}

Image denoising is one of the oldest problems in image processing and various denoising methods have been proposed over the past several decades, e.g., BM3D \citep{bm3d}, wavelet shrinkage \citep{SimAde96}, field of experts \citep{foe}, sparse-coding based approach \citep{mai09}, WNNM \citep{wnnm}, EPLL \citep{ZorWei11} and CSF \citep{csf}, etc. 

In this paper, we propose a new image denoiser, dubbed as Neural Affine Image Denoiser (Neural AIDE), which utilizes neural network in a novel way. The method is inspired by the recent work in discrete denoising \citep{MooMinLeeYoo16}, in which a novel ``pseudo-labels'' were devised to train a denoiser solely based on the noisy data. We extend the approach to the continuous-valued data case and devise a novel estimated loss function based on the noisy data that is an unbiased estimate of the true MSE. By investigating the devised estimated loss function we formulate to train a neural network to learn an \emph{affine} mapping that gets applied to a noisy pixel, based on its context. Such formulation enables
both supervised training of the network from the labeled training dataset and adaptive fine-tuning of the network parameters using the given noisy image subject to denoising. Our experimental results extensively show how we made subtle design choices in developing our algorithm.  Furthermore, we show that Neural AIDE significantly outperforms strong state-of-the-art baselines in the standard benchmark test datasets.

\section{Notations and Problem Setting }\label{sec:main results}


We denote $x^{n\times n}$ as the clean grascale image, and each pixel $x_i\in \{ 0,\ldots,255 \}$ is corrupted by an independent additive noise to result in a noisy pixel $Z_i$, \ie,
\be
Z_i = x_i + N_i, \ \ \ i=1, \ldots, n^2,\label{eq:noise_model}
\ee
where the continuous noise variables $N_i$'s are independent (not necessarily identically distributed nor Gaussian) over $i$ and $\E(N_i)=0, \mathbb{E}(N_i^2)=\sigma^2$ for all $i$. 
Moreover, As in the standard processing in grayscale image denoising, we normalize both $x_i$'s and $Z_i$'s with $255$ and treat them as real numbers. 
Importantly, following the \emph{universal} setting in discrete denoising \citep{Dude,MooMinLeeYoo16}, we treat the clean image $x^{n\times n}$ as an \emph{individual} image without any probabilistic model and only treat $Z^{n\times n}$ as random.

Generally, a denoiser can be denoted as $\hat{X}^{n\times n}=\{ \hat{X}_i(Z^{n\times n})\}_{i=1}^{n^2}$ denoting that each reconstruction at location $i$ is a function of the noisy image $Z^{n\times n}$. The standard loss function used for the grayscale image denoising to measure the denoising quality is the mean-squared error (MSE) denoted as 
\be
\Lb_{\hat{X}^{n\times n}}(x^{n\times n},Z^{n\times n})&=& \frac{1}{n^2}\sum_{i=1}^{n^2}\Lb\big(x_i,\hat{X}_i(Z^{n\times n})\big)
\ee
where $\Lb(x,\hat{x})=(x-\hat{x})^2$ is the per-symbol squared-error. Conventionally, the MSE is compared in the dB-scale using the Peak Signal-to-Noise-Ratio (PSNR) defined as $10\log_{10} (1/\Lb_{\hat{X}^{n\times n}}(x^{n\times n},Z^{n\times n}))$. 

\subsection{Estimated loss function for the affine denoiser}

In this paper, we consider the denoiser of the form
$
\hat{X}_i(Z^{n\times n}) = a(Z^{\backslash i}) \cdot Z_i +b(Z^{\backslash i})
$
for each $i$, in which $Z^{\backslash i}$ stands for the entire noisy image \emph{except} for $Z_i$.
Namely, the reconstruction at location $i$ has the \emph{affine} function form of the noisy symbol $Z_i$, but the slope and the intercept parameters, \ie, $a(Z^{\backslash i})$ and $b(Z^{\backslash i})$, of the affine function can be functions of the surrounding pixels. Hence, separete parameters can be learned from data for each location. Before presenting more concrete form of our denoiser, we first consider the following lemma.  
\begin{lemma}\label{lem:unbiased}
Consider a single-symbol case $Z=x+N$ with $\E(N)=0$ and $\E(N^2)=\sigma^2$, and suppose a single-symbol denoiser has the form of $\hat{X}(Z)=aZ+b$. Then, 
\be
\Ellb(Z,(a,b);\sigma^2) = (Z-(aZ+b))^2+2a\sigma^2\label{eq:est_loss}
\ee
is an unbiased estimate of $\E_x\Lb(x,\hat{X}(Z))+\sigma^2$, in which $\Lb(x,\hat{x})=(x-\hat{x})^2$ and $\E_x(\cdot)$ notation stands for the expectation over $Z$ given that the clean symbol is $x$.
\end{lemma}
\emph{Remark:} Note while the true MSE, $\Lb(x,\hat{X}(Z))$, can be evaluated only when the clean symbol $x$ is known, the estimated loss $\Ellb(Z,(a,b))$ can be evaluated soley with the noisy symbol $Z$, the affine mapping $(a,b)$ and the noisy variance $\sigma^2$. Thus, $\Ellb(Z,(a,b))$ plays a key role in adaptively learning the neural network-based affine denoiser as shown in the next section. 

\emph{Proof:}
By simple algebra, we have the following equalities:
\be
\E_x(x-\hat{X}(Z))^2 &=& \E_x (x^2+(aZ+b)^2-2x(aZ+b))\nonumber \\
 			         &=& \E_x (x^2+(aZ+b)^2-2ax^2-2bx) \label{eq:lem_1}\\
 			         &=& \E_x (Z^2-\sigma^2+(aZ+b)^2-2a(Z^2-\sigma^2)-2bZ)\label{eq:lem_2}\\
 			         &=& \E_x \Big(\big(Z-(aZ+b)\big)^2 +(2a-1)\sigma^2\Big)\label{eq:lem_3}\\
 			         &=& \E_x \Ellb(Z,(a,b);\sigma^2)-\sigma^2,\nonumber
\ee
in which (\ref{eq:lem_1}) follows from $\E_x(Z)=x$, (\ref{eq:lem_2}) follows from $\E_x(Z^2)=x^2+\sigma^2$ and replacing $x^2$ with $\E_x(Z^2-\sigma^2)$, and (\ref{eq:lem_3}) follows from simply rearranging the terms. Thus, we have the lemma. \ \qed

From Lemma \ref{lem:unbiased}, we can also show that for the denoisers of the form $
\hat{X}_i(Z^{n\times n}) = a(Z^{\backslash i}) \cdot Z_i +b(Z^{\backslash i}), 
$
\be
\E_{x_i}\Big(\Lb(x_i,\hat{X}_i(Z^{n\times n}))\big| Z^{\backslash i}\Big) = \E_{x_i}\Big( \Ellb(Z_i,(a(Z^{\backslash i}) ,b(Z^{\backslash i}));\sigma^2) | Z^{\backslash i} \Big) - \sigma^2\label{eq:cond_unbiased}
\ee
holds since $a(Z^{\backslash i})$ and $b(Z^{\backslash i})$ become constant given $Z^{\backslash i}$ and the noise is independent over $i$. The $\E_{x_i}(\cdot | Z^{\backslash i})$ in (\ref{eq:cond_unbiased}) stands for the conditional expectation of $Z_i$ given the clean symbol $x_i$ and the noisy symbols $Z^{\backslash i}$. Note the estimated loss function similar to (\ref{eq:est_loss}) has been also used to the filtering problem \citep{MooWei09b}.

\section{Neural AIDE: Neural Affine Image DEnoiser}

\subsection{Neural network-based affine denoiser}\label{subsec:affine denoiser}

Our proposing Neural Affine Image DEnoiser (Neural AIDE) considers the denoiser of the form 
\be
\hat{X}_i(Z^{n\times n}) = a(\mathbf{C}_{k\times k}^{\backslash i}) \cdot Z_i + b(\mathbf{C}_{k\times k}^{\backslash i}),\ \ \ i=1,\ldots,n\times n\label{eq:n_aide}
\ee
in which $\mathbf{C}_{k\times k}^{\backslash i}$ stands for the noisy image patch, or the \emph{context}, of size $k\times k$ surrounding $Z_i$ that does \emph{not} include $Z_i$. Thus, the patch has a \emph{hole} in the center. Then, we define a neural network 
\be
\mathbf{g}(\wb,\cdot):[0,1]^{k^2-1}\rightarrow \mathbb{R}_+^2\label{eq:nn_defn}
\ee
that takes the context $\mathbf{C}_{k\times k}^{\backslash i}$ as input and outputs the slope and intercept parameters $a(\mathbf{C}_{k\times k}^{\backslash i})$ and $b(\mathbf{C}_{k\times k}^{\backslash i})$ for each location $i$. We denote $\wb$ as the weight parameters of the neural network, which will be learned by the process described in the later sections. As it will get clear in our arguments below, the specific form of our denoiser in (\ref{eq:n_aide}) enables learning the parameters by both supervised learning with labelled training data and adaptive fine-tuning with the given noisy image. 

\begin{wrapfigure}{r}{0.2\textwidth}
\includegraphics[width=1\linewidth]{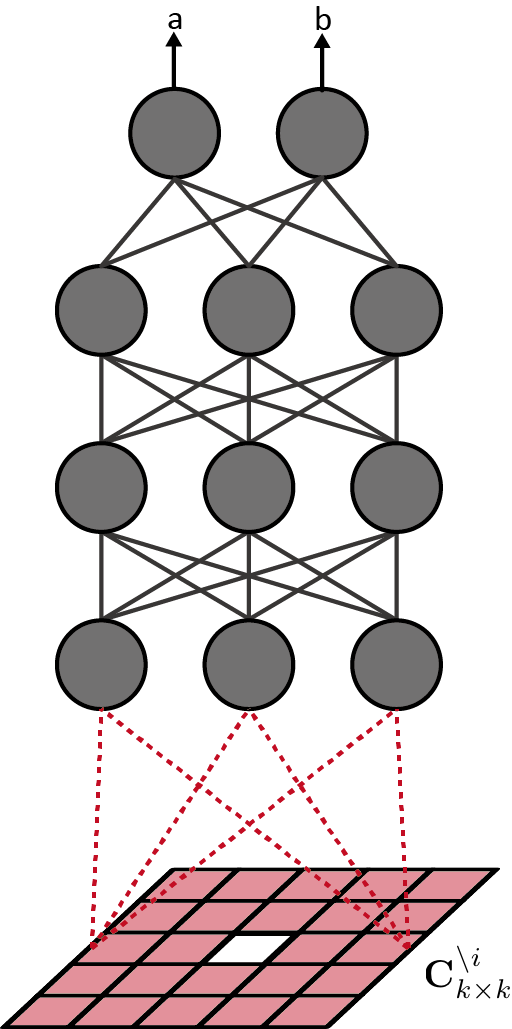} 
\caption{The architecture of Neural AIDE}
\label{fig:ber_and_est_ber_dna}
\end{wrapfigure}
Note in (\ref{eq:nn_defn}), we put a constraint that the slope and intercept of the affine function, \ie, the output of the network, should be nonnegative. While such constraint would appear apparent in our experimental results, it also makes an intuitive sense; the denoiser (\ref{eq:n_aide}) tries to estimate $x_i$ from $Z_i$, which are both in the interval $[0,1]$, hence, the nonnegative slope and intercept parameters should suffice. The nonnegativity constraint is realized in the neural network by applying 
\be
f(x) = \log (1+e^x)\label{eq:positive_activation}
\ee 
as the activation function for the final output layer of the neural network. The rest of the network architecture is the ordinary fully-connected neural network with ReLU activation functions, as depicted in Figure 1.




There are two sharp differences with our Neural AIDE and other neural network based denoisers, \eg, \citep{BurSchHar12,XieXuChe12}. First, the other schemes take the full noisy image patch (including the center location) as input to the network, and the network is trained to directly infer the corresponding clean image patches. In contrast, Neural AIDE is trained to first learn an affine \emph{mapping} based on the noisy image patch with a hole (\ie, the context of $Z_i$), then the learned mapping is applied to $Z_i$ to obtain the recostruction $\hat{X}_i$. Such difference enables the development of the estimated loss function in Lemma \ref{lem:unbiased} and the adaptive training process described in the next section. The principle of learning a mapping first and applying the mapping to the noisy symbol for denoising or filtering has been utilized in \citep{UFP06,MooWei09b,MooMinLeeYoo16}.
Second, unlike the other schemes, in which the patch-level reconstructions should somehow be aggregated to generate the final denoised image, Neural AIDE simply generates the final pixel-by-pixel reconstructions. Thus, there is no need for a step to aggregate multiple number of reconstructed patches, which simplifies the denoising step. 
Furthermore, since the neural network of Neural AIDE only has to estimate the two parameters of the affine mapping from each context, Neural AIDE can make much more efficient usage of the data with a simpler model compared to the networks in other schemes that need to estimate the full $k\times k$-patch, \eg, \cite{BurSchHar12}. 


\subsection{Adaptive training with noisy image}\label{subsec:adaptive training}

We first describe how the network parameters $\wb$ can be adaptively learned from the given noisy image $Z^{n\times n}$ without any additional labelled training data.
That is, by denoting each output element of the neural network $\mathbf{g}(\wb,\cdot)$ for the context $\Cb_{k\times k}^{\backslash i}$ as
\be
\mathbf{g}(\wb,\Cb_{k\times k}^{\backslash i})_1 \triangleq a(\Cb_{k\times k}^{\backslash i}) \ \ \text{and} \ \  \mathbf{g}(\wb,\Cb_{k\times k}^{\backslash i})_2 \triangleq b(\Cb_{k\times k}^{\backslash i}),\nonumber
\ee
we can define an objective function for the neural network to minimize as
\be
\mathcal{L}_{\text{adaptive}}(\wb, Z^{n\times n}) \triangleq \frac{1}{n^2}\sum_{i=1}^{n^2}\Ellb\Big(Z_i,(
\mathbf{g}(\wb,\Cb_{k\times k}^{\backslash i})_1,
\mathbf{g}(\wb,\Cb_{k\times k}^{\backslash i})_2);\sigma^2\Big)\label{eq:direct_denoising}
\ee
by using the estimated loss function $\Ellb(Z,(a,b);\sigma^2)$ defined in Lemma \ref{lem:unbiased}. The training process using (\ref{eq:direct_denoising}) is identical to the ordinary neural network learning, \ie, start with randomly initiallized $\wb$, then use backprogagation and variants of mini-batch SGD for updating the parameters.

The formulation (\ref{eq:direct_denoising}) may seem similar to training a neural network for a regression problem; namely, $\{(\Cb_{k\times k}^{\backslash i},Z_i)\}_{i=1}^{n^2}$, which are solely obtained from the noisy image $Z^{n\times n}$, can be analogously thought of as the input-target label pairs for the supervised regression. But, unlike regression, which tries to directly learn a mapping from input to the target label, our network learns the affine mapping for each context and apply it to $Z_i$ to estimate the unobserved clean symbol $x_i$. 
The fact that (\ref{eq:direct_denoising}) only depends on the given noisy image $Z^{n\times n}$ (and the assumed $\sigma^2$) makes the learning adaptive. 

The rationale behind using $\Ellb(Z,(a,b);\sigma^2)$ in (\ref{eq:direct_denoising}) is the following; as shown in (\ref{eq:cond_unbiased}), the estimated loss is an unbiased estimate of the true expected squared-error given the context $\Cb_{k\times k}^{\backslash i}$. Therefore, minimizing (\ref{eq:direct_denoising}) may result in the network that produces the slope and intercept parameters that minimize the true MSE for the reconstrunctions of the corresponding affine mappings. This formulation of training neural network parameters solely based on the noisy data is inspired by the recent work in discrete denoising \citep{MooMinLeeYoo16}. 

Once the training is done, we can then denoise the very noisy image $Z^{n\times n}$ used for training by applying the affine mapping at each location as (\ref{eq:n_aide}). That is, by denoting $\wb^*$ as the learned parameter by minimizing (\ref{eq:direct_denoising}), the reconstruction at location $i$ by Neural AIDE becomes
\be
\hat{X}_{i,\text{Neural AIDE}}(Z^{n\times n}) = \mathbf{g}(\wb^*,\Cb_{k\times k}^{\backslash i})_1\cdot Z_i + \mathbf{g}(\wb^*,\Cb_{k\times k}^{\backslash i})_2.\label{eq:n_aide_denoise}
\ee





\subsection{Supervised training and adaptive fine-tuning}\label{subsec:learning}

While the formulation in (\ref{eq:direct_denoising}) gives an effective way of adaptively training a denoiser based on the given noisy image $Z^{n\times n}$, the specific form of the denoiser in (\ref{eq:n_aide}) makes it possible to carry out the supervised pre-training of $\wb$ before the adaptive training step. That is, we can collect abundant clean images, $\tilde{x}^{n\times n}$, from the various image sources (\eg, World Wide Web) and corrupt them with the assumed additive noise with variance $\sigma^2$ in (\ref{eq:noise_model}) to generate the correspoding noisy images, $\tilde{Z}^{n\times n}$, and the labelled training data of size $N$,
\be
\mathcal{D} = \{ (\tilde{x}_i, \tilde{\Cb}_{i,k\times k})\}_{i=1}^{N}. \label{eq:sup_data}
\ee
In (\ref{eq:sup_data}), $\tilde{\Cb}_{i,k\times k}$ stands for the noisy image patch of size $k\times k$ at location $i$ that \emph{includes} the noisy symbol $\tilde{Z}_i$, and $\tilde{x}_i$ is the clean symbol that correspond to $\tilde{Z}_i$. Now, the subtle point is that, unlike the usual supervised learning that may directly learn a mapping from $\tilde{\Cb}_{i,k\times k}$ to $\tilde{x}_i$, we remain in using the neural network defined in (\ref{eq:nn_defn}) and learn $\wb$ by minimizing 
\be
\mathcal{L}_{\text{supervised}}(\wb, \mathcal{D}) \triangleq \frac{1}{N}\sum_{i=1}^{N}\Lb\Big(\tilde{x}_i,
\mathbf{g}(\wb,\tilde{\Cb}_{k\times k}^{\backslash i})_1\cdot \tilde{Z}_i+
\mathbf{g}(\wb,\tilde{\Cb}_{k\times k}^{\backslash i})_2\Big).\label{eq:supervised_denoising}
\ee
Note $\Lb(x,\hat{x})=(x-\hat{x})^2$ as before. The training process of minimizing (\ref{eq:supervised_denoising}) is again done by the usual backpropagation and the variants of mini-batch SGD. 

Once the objective function (\ref{eq:supervised_denoising}) converges after sufficient iteration of weight updates, we denote the converged parameter as $\tilde{\wb}$. Then, for a given noisy image to denoise, $Z^{n\times n}$, we can further update $\tilde{\wb}$ adaptively for $Z^{n\times n}$ by minimizing $\mathcal{L}_{\text{adaptive}}(\wb, Z^{n\times n})$ in (\ref{eq:direct_denoising}) starting from $\tilde{\wb}$. That is, we adaptively \emph{fine-tune} $\tilde{\wb}$ until $\mathcal{L}_{\text{adaptive}}(\wb, Z^{n\times n})$ converges, then denoise $Z^{n\times n}$ with the converged parameter as (\ref{eq:n_aide_denoise}). This capability of adaptively fine-tuning the supervised trained weight parameter is the unique characteristic of Neural AIDE that differentiates it from other neural network-based denoisers.

\section{Experimental Results}\label{sec:experiments}
    
   

We compared the denoising performance of the proposed Neural AIDE with several state-of-the-art denoising methods, including BM3D \citep{bm3d}, MLP \citep{BurSchHar12}, EPLL \citep{ZorWei11}, WNNM \citep{wnnm} and CSF \citep{csf}. 

\subsection{Data and experimental setup}

For the supervised training, we generated the labelled training set using 2000 images available in public datasets. Out of 2000 images, 300 images are taken from train/validation set in the Berkeley Segmentation Dataset and the remaining 1700 images are taken from Pascal VOC 2012 Dataset. For the Pascal VOC images, we resized them to match the resolution of the Berkeley Segmentation Dataset \citep{berkeley}, $481\times 321$. We corrupted the images with additive Gaussian noise and tested with multiple noise levels, namely, $\sigma = 5,10,15,20,25$. That is, we built separate training set of size 2000 for each noise level. The total number of training data points (\ie, $N$ in (\ref{eq:sup_data})) in each dataset was thus about 308 million. We evaluated the performance of the denoisers with 11 standard test images, \ie, $\{$Barbara, Boat, C.man Couple, F.print, Hill, House, Lena, Man, Montage and Peppers$\}$, and 68 standard Berkeley images \citep{foe}. 

Our network had 9 fully connected layers with 512 nodes in each layer, which showed the best result among a few tried models \footnote{The difference among the models were not huge.}. ReLU was used as activation functions, and we used Adam \citep{KinBa15} as the optimizer to train the network. For the supervised training, we trained the network up to 50 epochs and halved the learning rate every 10 epochs starting from $10^{-4}$. For the adaptive fine-tuning, we also trained up to 50 epochs and halved the learning rate every 20 epochs starting from $10^{-5}$. We did not use any regularization methods while training. Moreover, for the context data, $\Cb_{k\times k}^{\backslash i}$, we subtracted $0.5$ from the values to make the input to the network get centered around 0. (Note $Z_i$ that the affine mappping gets applied to in (\ref{eq:n_aide_denoise}) still is in the original scale.)

For all our experiments, we used Keras (version 1.2.2) with Tensorflow (version 0.11.0) backend and NVIDIA's GPU (GeForce GTX1080) with CUDA library version 8.0.




\subsection{Training Neural AIDE}
In this section, we systematically show the reasoning behind choosing the context size $k$, the empirical justification of the nonnegative contraint on the outputs of $\mathbf{g}(\wb,\cdot)$ and the validity of the combination of the supervised pre-training with adaptive fine-tuning.

\subsubsection{Adaptive training with noisy image}
\begin{figure*}[h]
    \centering
     \subfigure[Adaptive training (random initialization)]{\label{fig:rand_init}
    \includegraphics[width=0.4\textwidth]{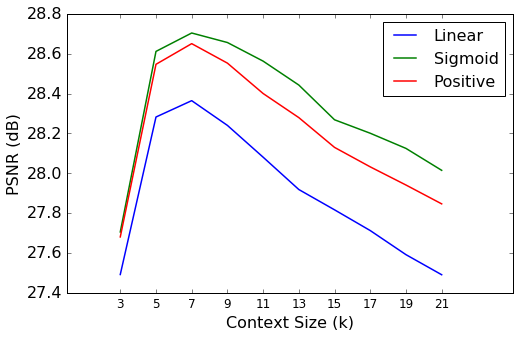}}
    \hspace{.5in}
     \subfigure[Supervised training (300 training images)]{\label{fig:k_size}
    \includegraphics[width=0.4\textwidth]{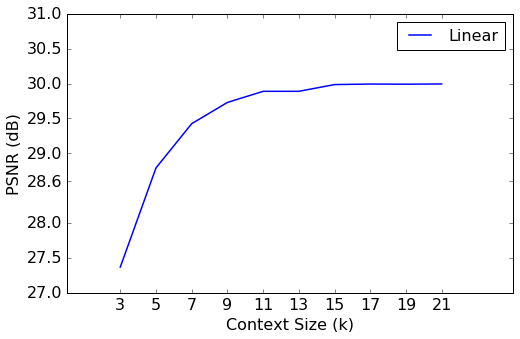}}
         \caption{Adaptive and supervised training results on the standard 11 test images ($\sigma=25$)}\label{fig:prelim}
\end{figure*}

We first carried out the adaptive training solely with the given noisy image as described in Section \ref{subsec:adaptive training}. That is, for each given noisy image, we randomly initialized the weight parameters of the neural network and trained with the objective function (\ref{eq:direct_denoising}). After training, the image was denoised as (\ref{eq:n_aide_denoise}). Figure \ref{fig:rand_init} shows the PSNR results on the standard 11 test images with varying $k$ values and output activation functions, \ie, Linear ($f(x)=x$), Positive ($f(x)=\log(1+e^x)$ in (\ref{eq:positive_activation})) and Sigmoid ($f(x)=1/(1+e^{-x})$). The noise level was $\sigma=25$.

From the figure, we can see that the adaptive training alone can still result in a decent denoiser, although some PSNR gap exists compared to the state-of-the-arts as shown in Table \ref{images_11}. We see that $k=7$ tend to be the best context size for adaptive training. Moreover, the choice of the output activation functions turns out to be important, and more discussion is given on the activation function in the next section.

\subsubsection{Supervised training and adaptive fine-tuning}




Since the limitation of the adaptive training alone was apparent, we then carried out the supervised training in Section \ref{subsec:learning}. That is, we took the 300 images from the Berkeley Segmentation Dataset and trained the network with varying $k$ values as shown in Figure \ref{fig:k_size}. Denoising of the noisy image was done identically as before by applying the learned affine mapping to each noisy pixel. Note in this case, we only carried out the experiments with the Linear activation function. We can see that the supervised training can result in a much higher PSNR values than the adaptive training, already very close to the state-of-the-arts. Also, the performance seems to get saturated around $k=17$, so in all our experiments below, we used $k=17$.

\begin{figure*}[h]
    \centering
    \subfigure[F.print(Lin.,\texttt{s})]{\label{fig:f_lin_s}
    \includegraphics[width=0.2\textwidth]{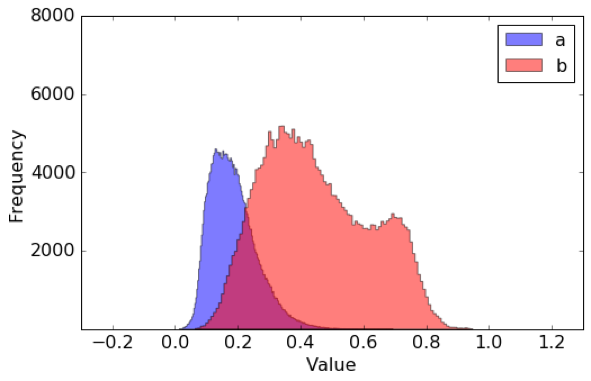}}
    \subfigure[F.print(Lin.,\texttt{ft})]{\label{fig:f_lin_ft}
    \includegraphics[width=0.2\textwidth]{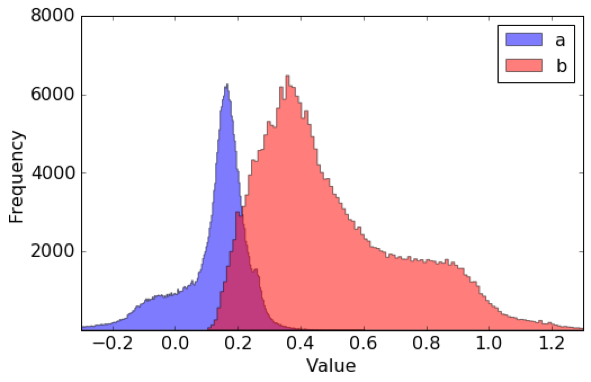}}
    \subfigure[F.print(Pos.,\texttt{s})]{\label{fig:f_pos_s}
    \includegraphics[width=0.2\textwidth]{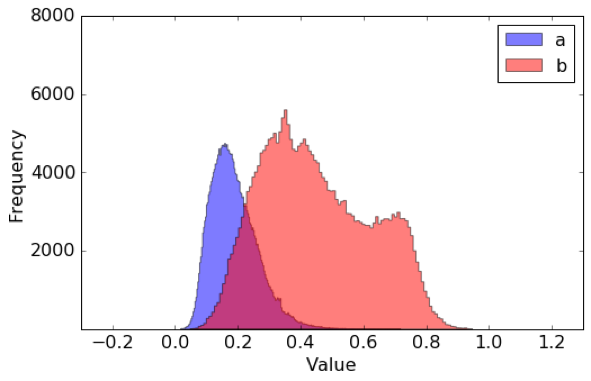}}
    \subfigure[F.print(Pos.,\texttt{ft})]{\label{fig:f_pos_ft}
    \includegraphics[width=0.2\textwidth]{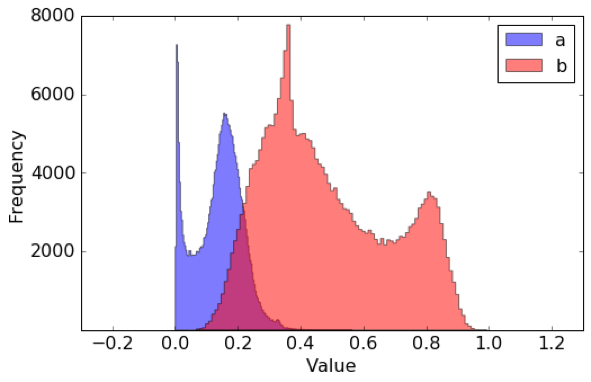}}
    \subfigure[Montage(Lin.,\texttt{s})]{\label{fig:m_lin_s}
    \includegraphics[width=0.2\textwidth]{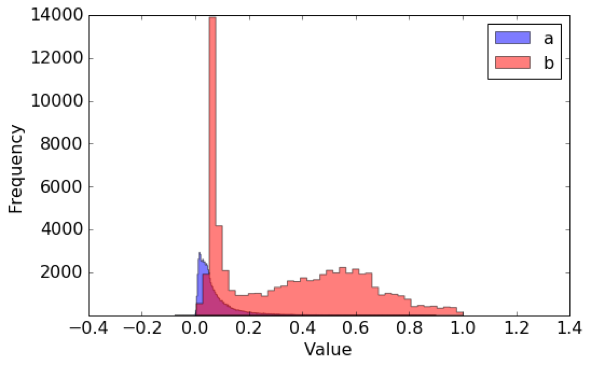}}
    \subfigure[Montage(Lin.,\texttt{ft})]{\label{fig:m_lin_ft}
    \includegraphics[width=0.2\textwidth]{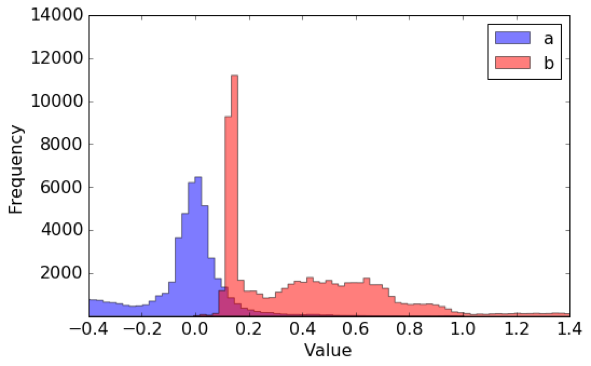}}
    \subfigure[Montage(Pos.,\texttt{s})]{\label{fig:m_pos_s}
    \includegraphics[width=0.2\textwidth]{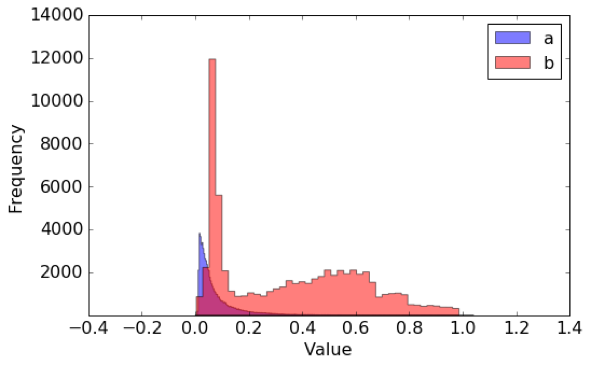}}
    \subfigure[Montage(Pos.,\texttt{ft})]{\label{fig:m_pos_ft}
    \includegraphics[width=0.2\textwidth]{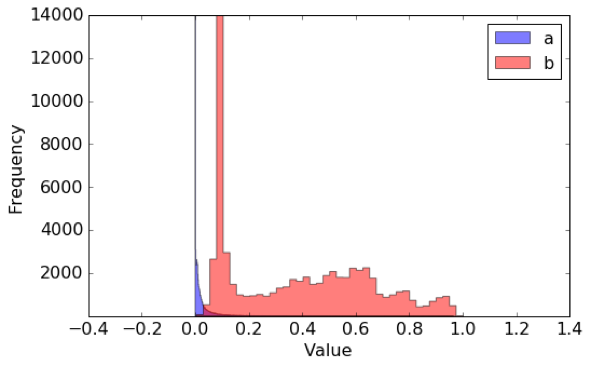}}
      \subfigure[PSNR values during adaptive fine-tuning.]{\label{fig:ft_curve}
    \includegraphics[width=0.4\textwidth]{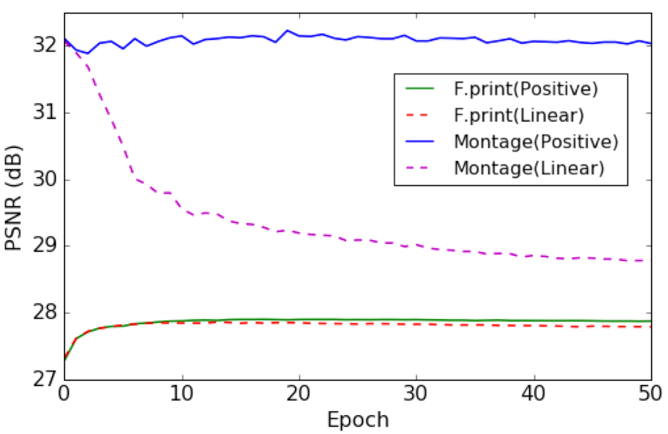}}
    \caption{(a-h) Distribution of $a$ and $b$ values for F.print and Montage after supervised training (\texttt{s}) and fine-tuning (\texttt{ft}) for Linear (Lin.) and Positive (Pos.) activation functions. The distributions obtained for fine-tuning are from the models at 50 epoch. (i) PSNR values during fine-tuning.}\label{fig:ab_dist}
\end{figure*}

Encouraged by this result, we moved on to adaptively fine-tuning the weight parameters by minimizing the objective function (\ref{eq:direct_denoising}) for each image initialized with the parameters learned by supervised learning. This is when the subtle issue regarding the activation function we describe below comes up. In Figure \ref{fig:ab_dist}, we trained supervised learning models with Linear and Positive output activation functions using 800 images for $\sigma=25$, then adaptively fine-tuned the parameters for given noisy image (F.print and Montage image). Figure \ref{fig:f_lin_s}-\ref{fig:f_pos_ft} show the distributions of the slope ($a$) and intercept ($b$) paramters that each model outputs for the given image, and \ref{fig:ft_curve} shows the change of PSNR value in the process of adaptive fine-tuning. 
From Figure \ref{fig:f_lin_s} and \ref{fig:m_lin_s}, we can see that when trained with supervised learning with Linear output activation function, the values of $a$ and $b$ all lie in the interval $[0,1]$. However, when fine-tuned for each image, Figure \ref{fig:f_lin_ft} and \ref{fig:m_lin_ft} show that many negative $a$ values are produced for the Linear activation. This can be readily seen by examining the form of $\Ellb(Z,(a,b);\sigma^2)$ in (\ref{eq:est_loss}), which does not hinder $a$ from having negative values when there is no constraint. As shown in Figure \ref{fig:ft_curve}, such negative $a$ values for the affine mapping sometime does not have big effect on the fine-tuning process and the final denoising performance as in the case of F.print, in which the PSNR increases significantly from the supervised model by fine-tuning. However, as in the case of Montage in Figure \ref{fig:ft_curve}, we suspect such negative $a$ values sometimes hurt the denoising performance greatly. In contrast, when we put the nonnegativitiy contstraint on $a$ and $b$ in the neural network, we observe a stable fine-tuning process, as is observed in Figure \ref{fig:f_pos_ft}, \ref{fig:m_pos_ft} and \ref{fig:ft_curve}. Thus, the results of Neural AIDE from now on all uses the positive activation function. \footnote{We also tested with the sigmoid activation and the result was more or less the same.}


\begin{figure*}[h]
    \centering
    \subfigure[PSNR]{\label{fig:psnr}
    \includegraphics[width=0.48\textwidth]{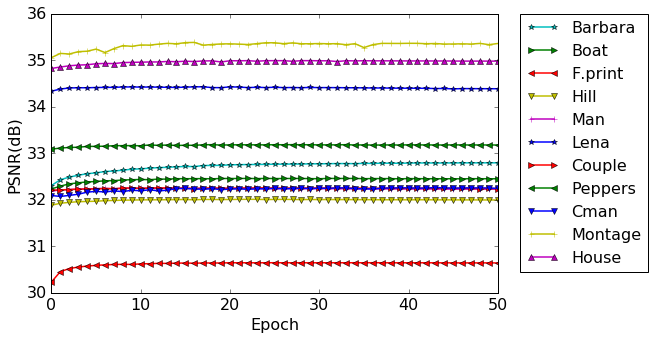}}
    \subfigure[Objective function (\ref{eq:direct_denoising})]{\label{fig:learning}
    \includegraphics[width=0.48\textwidth]{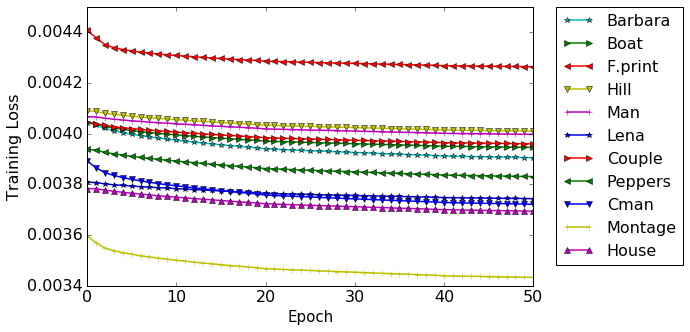}}
    \caption{PSNR and objective function value during fine-tuning for the standard 11 images ($\sigma=15$)}\label{fig:psnr_obj}
\end{figure*}

Figure \ref{fig:psnr_obj} shows the adaptive fine-tuning process of the standard 11 images for $\sigma=15$. The supervised model was trained with the full training set of 2000 images. From the figures, we can see that the learning is done appropriately and the PSNR does improve with fine-tuning. 

\newpage

\subsection{Quantitative evaluation}
\subsubsection{Standard 11 images}

Table 1 summarizes our denoising results compared to the recent state-of-the-arts on the standard 11 images for various noise levels. We show both mean and standard deviation of PSNR values. For the baseline methods, we downloaded the codes from the authors' webpages and ran the code on the noisy images, thus, the numbers can be compared fairly. (MLP and CSF$^{5}_{7\times 7}$ could run only on selected noise levels.) N-AIDE$_\texttt{S}$ stands for the Neural AIDE that is only supervised trained (with 2000 images). N-AIDE$_\texttt{fB}$ and N-AIDE$_\texttt{fH}$ are fine-tuned models after supervised learning; N-AIDE$_\texttt{fB}$ is the best model (in terms of epoch) chosen based on PSNR (thus, not practical) and N-AIDE$_\texttt{fH}$ is the model that is chosen with a heuristic rule - \ie, stop fine-tuning when the training loss becomes smaller than $\sigma^2$, otherwise fine-tune until 50 epochs. 

From the table, we can see that N-AIDE$_\texttt{fH}$ significantly outperforms all other baselines on average except for WNNM. The difference of mean PSNR between WNNM and N-AIDE$_\texttt{fH}$ is almost negligible and N-AIDE$_\texttt{fH}$ tend to have smaller variance in terms of PSNR than WNNM. By comparing N-AIDE$_\texttt{S}$ and N-AIDE$_\texttt{fH}$, we can definitely see that adaptive fine-tuning is effective. Also, when the noise level is low, the improvement gets larger. Furthermore, by comparing N-AIDE$_\texttt{S}$ with MLP, which is another neural network based denoiser and uses much more data points (362 million exmample) and larger model, we can confirm that our model more efficiently uses the data. 


\begin{table}[H]
\centering
    \begin{tabular}{| c|| c || c| c|c|c|c|c|c|c|}
    \hline
    $\sigma$ & \texttt{PSNR} & BM3D & MLP & EPLL & WNNM  & CSF$^{5}_{7\times 7}$ & N-AIDE$_\texttt{s}$ & N-AIDE$_\texttt{fB}$& N-AIDE$_\texttt{fH}$\\ \hline
        \hline
   \multirow{2}{*}{5} & Mean & 38.24& - &  37.88& \textbf{38.43}  &  -& 38.14 &38.44 &\textbf{38.44}\\ \cline{2-10}
                        & Std &  1.24 & - &  1.07 &  1.28 & - & 1.17& 1.18&1.18\\ \cline{2-10} \hline
    \multirow{2}{*}{10} & Mean & 34.71 &34.45  &34.27  & \textbf{34.95} &  - &34.66 & 34.92&\textbf{34.91}\\ \cline{2-10}
                        & Std &  1.37& 1.12 &1.18  & 1.42   & - &1.31 & 1.33&1.33\\ \cline{2-10} \hline
    \multirow{2}{*}{15} & Mean & 32.76 & - & 32.29 & \textbf{32.99} & 32.40  & 32.77& 32.97&\textbf{32.96}\\ \cline{2-10}
                        & Std & 1.48 & - & 1.35 &  1.54  & 1.27 & 1.44&1.42 &1.42\\ \cline{2-10} \hline
    \multirow{2}{*}{20} & Mean & 31.43 &  -& 30.90 & \textbf{31.59} &  - & 31.38& 31.58&\textbf{31.55}\\ \cline{2-10}
                        & Std & 1.50 & - & 1.34 & 1.57 & - & 1.50&1..46 &1.44\\ \cline{2-10} \hline
    \multirow{2}{*}{25} & Mean &  30.40& 30.24 & 29.81 & \textbf{30.51} & 29.93  & 30.36& 30.51&\textbf{30.47}\\ \cline{2-10}
                        & Std & 1.51 & 1.43 & 1.38 & 1.56 & 1.41 & 1.53&1.45 &1.46\\ \cline{2-10} \hline                                       
     \end{tabular}
    \vspace{.1in}\caption{PSNR comparsions on the 11 standard benchmark images for $\sigma=5,10,15,20,25$.  }
    \label{images_11}
\end{table}

\begin{figure*}[h]
    \centering
    \subfigure[Competitive comparison ]{\label{fig:competitive}
    \includegraphics[width=0.35\textwidth]{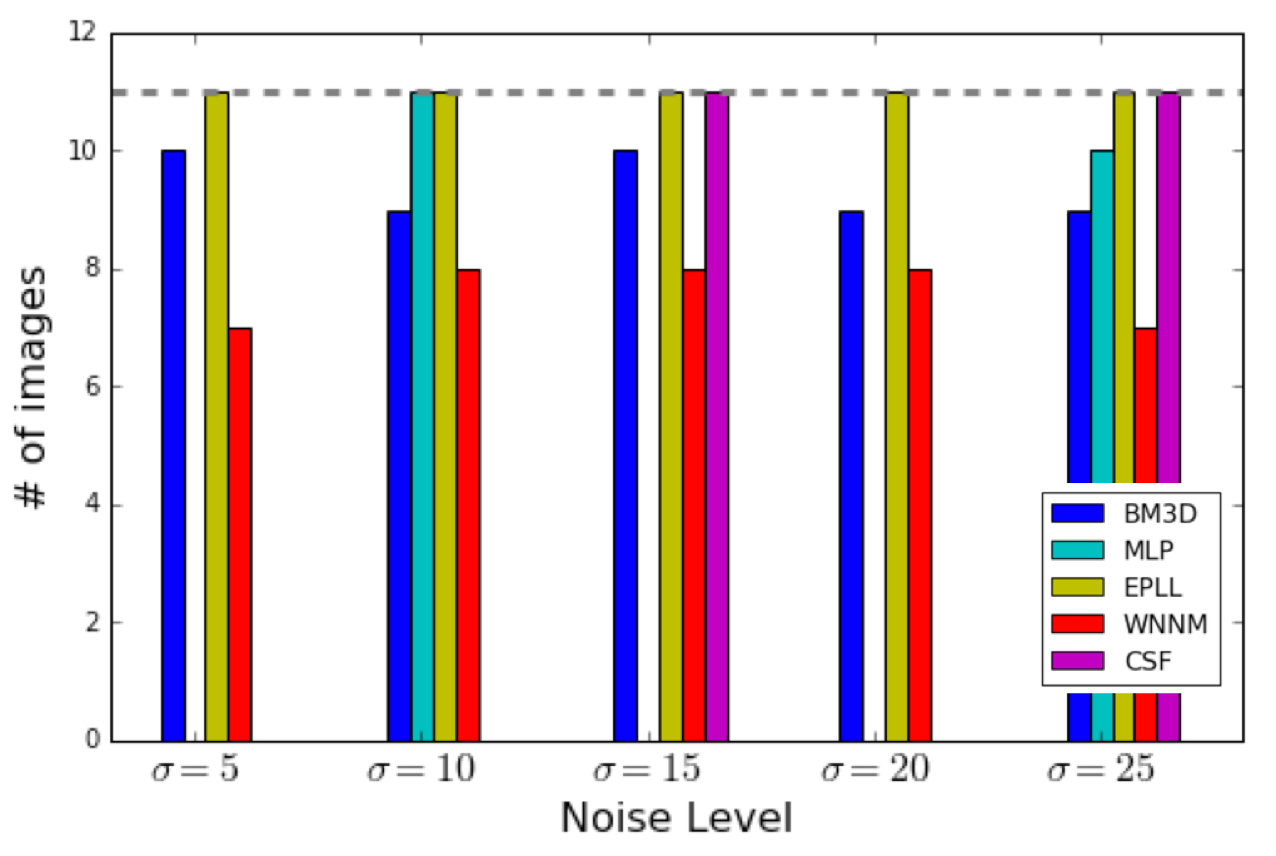}}
    \subfigure[PSNR of N-AIDE$_\texttt{s}$]{\label{fig:mismatch_supervised}
    \includegraphics[width=0.3\textwidth]{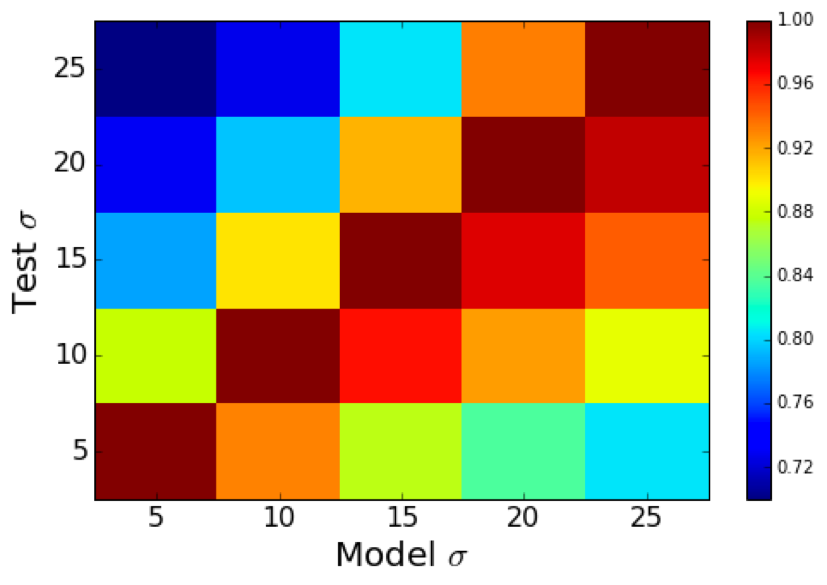}}
    \subfigure[PSNR of N-AIDE$_\texttt{fH}$ ]{\label{fig:mismatch_fine_tune}
    \includegraphics[width=0.3\textwidth]{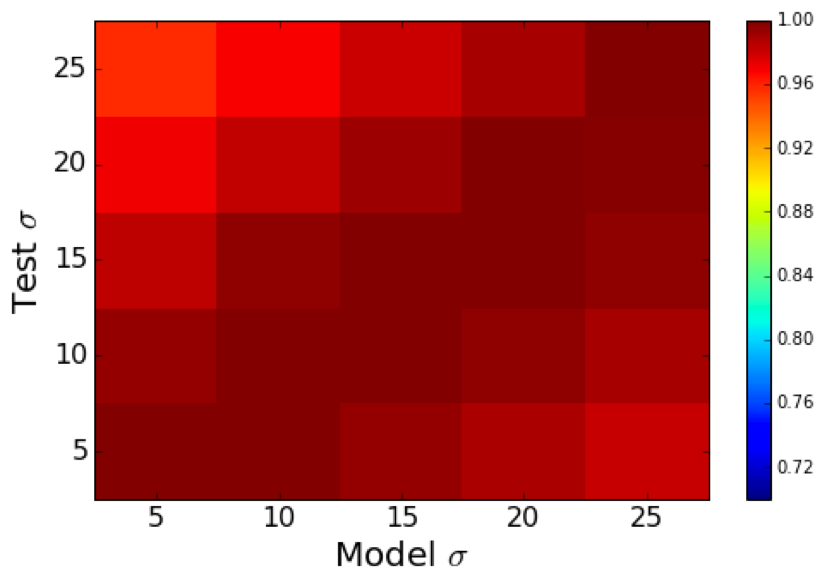}}   
    \caption{(a) Competitive comparison of  N-AIDE$_\texttt{fH}$ with baselines (b) PSNR of mismatched N-AIDE$_\texttt{s}$ (c) PSNR of N-AIDE$_\texttt{fH}$ with mismatched N-AIDE$_\texttt{s}$ but fine-tuned with correct $\sigma$}
\end{figure*}

Figure \ref{fig:competitive} shows the competitive comparison between N-AIDE$_\texttt{fH}$ and the baselines. That is, the figure plots the number of images of which the PSNR of N-AIDE$_\texttt{fH}$ is better than the baseline methods. We can see that our method mostly outperforms all baselines competitively, including WNNM. 

One of the main drawbacks of MLP \cite{BurSchHar12} is that the neural networks have to be trained separately for all noise levels and the mismatch of $\sigma$ significantly hurts the denoising performance. While the supervised training of Neural AIDE is also done in the similar way, Figure \ref{fig:mismatch_supervised}-\ref{fig:mismatch_fine_tune} show that the adaptive fine-tuning can be very effective in overcoming such limitation. Figure \ref{fig:mismatch_supervised} shows the PSNR results of the mismatched N-AIDE$_\texttt{s}$ models before fine-tuning. Each row is normalized with the PSNR of the matched case, \ie, the diagonal element, and the PSNR values are color-coded. We clearly see the sensitivity of PSNR in the mismatch of $\sigma$ as the off-diagonal values show significant gaps compared to the diagonal values in each row. On the other hand, Figure \ref{fig:mismatch_fine_tune} shows the PSNR values of N-AIDE$_\texttt{fH}$'s that have mismatched supervised models but are adaptively fine-tuned with the correct $\sigma$'s. We can clearly see that the PSNR gaps of the mismatched supervised models can be significantly closed by adaptive fine-tuning, which gives a significant edge over MLP in \citep{BurSchHar12}. 



\subsubsection{Standard 68 Berkeley images}

Table \ref{images_berkeley} shows the PSNR results on the 68 standard Berkeley images from \citep{foe}. We can clear see that N-AIDE$_\texttt{fH}$ again outperforms the baseline state-of-the-art methods, including WNNM, with significant margins.
\begin{table}[H]
\centering
    \begin{tabular}{| c||c|c|c|c|c|c|c|}
    \hline
    $\sigma$ &  MLP & EPLL & WNNM  & CSF$^{5}_{7\times 7}$  &N-AIDE$_\texttt{s}$ & N-AIDE$_\texttt{fB}$& N-AIDE$_\texttt{fH}$\\ \hline
        \hline
   \multirow{1}{*}{5}   &  -    &  37.50    & 37.71 &  -    &  37.72 & 37.82  & \textbf{37.79}\\ 
                       \hline
    \multirow{1}{*}{10} & 33.41 &  33.32    & 33.48 &  -    & 33.62& 33.71  &\textbf{33.66}\\ 
                       \hline
    \multirow{1}{*}{15} &  -    & 31.09     & 31.18 & 31.10 &  31.45 & 31.52  &\textbf{31.47}\\ 
                         \hline
    \multirow{1}{*}{20} &   -    & 29.60    &  29.63     &  -    & 29.98  &  30.05 &\textbf{30.00}\\ 
                        \hline
    \multirow{1}{*}{25}  & 28.73 & 28.47    &  28.46     & 28.41 &  28.93 & 28.97  &\textbf{28.90}\\ 
                         \hline                                       
     \end{tabular}
    \vspace{.1in}\caption{PSNR comparisons on the 68 standard Berkeley images. }
    \label{images_berkeley}
\end{table}

\section{Concluding remarks}

We devised a novel neural network based image denoiser, Neural AIDE. The algorithm is devised with a different principle from the other state-of-the-art methods. As a result, we show that a very simple adaptive affine model, which Neural AIDE learns differently for each pixel, can significantly outperform many strong baselines. Also, the adaptive fine-tuning of Neural AIDE can successfully overcome the $\sigma$ mismatch problem, which is a serious drawback of other neural network based methods. 

As a future work, we would like to more thoroughly carry out the experiments in even noisier regime. Also, since our algorithm does not require the noise to be Gaussian (only the additivity of the noise and $\sigma^2$ are assumed), we would try to other types of noise, \eg, Laplacian noise. Furthermore, extending our framework to non-additive noise such as multiplicative noise would be another interesting direction. Finally, theoretical anayses of our method based on information theory and learning theory would be another direction worth pursuing.

\newpage
\bibliographystyle{unsrtnat}

\begin{thebibliography}{15}
\providecommand{\natexlab}[1]{#1}
\providecommand{\url}[1]{\texttt{#1}}
\expandafter\ifx\csname urlstyle\endcsname\relax
  \providecommand{\doi}[1]{doi: #1}\else
  \providecommand{\doi}{doi: \begingroup \urlstyle{rm}\Url}\fi

\bibitem[Dabov et~al.(2007)Dabov, Foi, Katkovnik, and Egiazarian]{bm3d}
K.~Dabov, A.~Foi, V.~Katkovnik, and K.~Egiazarian.
\newblock Image denoising by sparse 3-d transform-domain collaborative
  filtering.
\newblock \emph{IEEE Trans. Image Processing}, 16\penalty0 (8):\penalty0
  2080--2095, 2007.

\bibitem[Simoncelli and Adelson(1996)]{SimAde96}
E.P. Simoncelli and E.H. Adelson.
\newblock Noise removal via bayesian wavelet coring.
\newblock In \emph{ICIP}, 1996.

\bibitem[Roth and Black(2009)]{foe}
S.~Roth and M.J Black.
\newblock Field of experts.
\newblock \emph{IJCV}, 82\penalty0 (2):\penalty0 205--229, 2009.

\bibitem[Mairal et~al.(2009)Mairal, Bach, Ponce, Sapiro, and Zisserman]{mai09}
J.~Mairal, F.~Bach, J.~Ponce, G.~Sapiro, and A.~Zisserman.
\newblock Non-local sparse models for image restoration.
\newblock In \emph{ICCV}, 2009.

\bibitem[Gu et~al.(2014)Gu, Zhang, Zuo, and Feng]{wnnm}
S.~Gu, L.~Zhang, W.~Zuo, and X.~Feng.
\newblock Weighted nuclear norm minimization with applicaitons to image
  denoising.
\newblock In \emph{CVPR}, 2014.

\bibitem[Zoran and Weiss(2011)]{ZorWei11}
D.~Zoran and Y.~Weiss.
\newblock From learning models of natural image patches to whole image
  restoration.
\newblock In \emph{ICCV}, 2011.

\bibitem[Schmidt and Roth(2014)]{csf}
U.~Schmidt and S.~Roth.
\newblock Shrinkage fields for effective image restoration.
\newblock In \emph{CVPR}, 2014.

\bibitem[Moon et~al.(2016)Moon, Min, Lee, and Yoon]{MooMinLeeYoo16}
T.~Moon, S.~Min, B.~Lee, and S.~Yoon.
\newblock Neural universal discrete denosier.
\newblock In \emph{NIPS}, 2016.

\bibitem[Weissman et~al.(2005)Weissman, Ordentlich, Seroussi, Verdu, and
  Weinberger]{Dude}
T.~Weissman, E.~Ordentlich, G.~Seroussi, S.~Verdu, and M.~Weinberger.
\newblock Universal discrete denoising: {K}nown channel.
\newblock \emph{{IEEE} Trans. Inform. Theory}, 51\penalty0 (1):\penalty0 5--28,
  2005.

\bibitem[Moon and Weissman(2009)]{MooWei09b}
T.~Moon and T.~Weissman.
\newblock Universal {FIR MMSE} filtering.
\newblock \emph{{IEEE} Transactions on Signal Processing}, 57\penalty0
  (3):\penalty0 1068--1083, 2009.

\bibitem[Burger et~al.(2012)Burger, Schuler, and Harmeling]{BurSchHar12}
H.~Burger, C.~Schuler, and S.~Harmeling.
\newblock Image denoising: {C}an plain neural networks compete with {BM3D}?
\newblock In \emph{CVPR}, 2012.

\bibitem[Xie et~al.(2012)Xie, Xu, and Chen]{XieXuChe12}
J.~Xie, L.~Xu, and E.~Chen.
\newblock Image denoising and inpainting with deep neural networks.
\newblock In \emph{NIPS}, 2012.

\bibitem[Weissman et~al.(2007)Weissman, Ordentlich, Weinberger, Somekh-Baruch,
  and Merhav]{UFP06}
T.~Weissman, E.~Ordentlich, M.~Weinberger, A.~Somekh-Baruch, and N.~Merhav.
\newblock Universal filtering via prediction.
\newblock \emph{{IEEE} Trans. Inform. Theory}, 53\penalty0 (4):\penalty0
  1253--1264, 2007.

\bibitem[Martin et~al.(2001)Martin, Fowlkes, Tal, and Malik]{berkeley}
D.~Martin, C.~Fowlkes, D.~Tal, and J.~Malik.
\newblock A database of human segmented natural images and its application to
  evaluating segmentation algorithms and measuring ecological statistics.
\newblock In \emph{ICCV}, 2001.

\bibitem[Kingma and Ba(2015)]{KinBa15}
D.~Kingma and J.~Ba.
\newblock Adam: A method for stochastic optimization.
\newblock In \emph{ICLR}, 2015.

\end{thebibliography}

\end{document}